\documentclass[conference,a4paper]{IEEEtran}
\IEEEoverridecommandlockouts

\usepackage[hidelinks]{hyperref}
\usepackage[cmex10]{amsmath}
\usepackage{amssymb,amsfonts}
\usepackage{dblfloatfix}

\usepackage[ruled,vlined]{algorithm2e}
\usepackage{graphicx}
\graphicspath{{Figures/PDF/}{Figures/PNG/}}

\usepackage{booktabs}
\usepackage{siunitx}
\usepackage[numbers,compress]{natbib}
\usepackage{texnames}
\usepackage{bm,bbm}
\usepackage{orcidlink}

\usepackage{arydshln}
\usepackage{multirow}
\usepackage{makecell}
\usepackage{tabularx}
\usepackage[table]{xcolor} 
\usepackage{array}
\usepackage{bigdelim} 

\begin{document}

\title{
\uppercase{Reconstructing Multi-Decadal Forest Disturbances: A Spatio-Temporal Transformer Approach}
}

\author{	\IEEEauthorblockN{Linus Scheibenreif\orcidlink{0000-0001-5580-8910}${}^{*}$\thanks{${}^{*}$Work done at Google DeepMind.}}
	\IEEEauthorblockA{\textit{ETH Zurich}\\
		Zurich, Switzerland\\
		lscheibenrei@ethz.ch}
	\and
	\IEEEauthorblockN{Anton Raichuk\orcidlink{0009-0000-4700-0815}}
	\IEEEauthorblockA{\textit{Google DeepMind}\\
		Zurich, Switzerland\\
		raveman@google.com}
	\and
	\IEEEauthorblockN{Maxim Neumann\orcidlink{0000-0002-1967-318X}}
	\IEEEauthorblockA{\textit{Google DeepMind}\\
		Zurich, Switzerland\\
		maximneumann@google.com}
}

\maketitle
\vspace*{-2em}

\begin{abstract}
%
Accurate monitoring of forest disturbances is essential for understanding carbon dynamics and land management, yet traditional approaches typically rely on pixel-wise analysis of satellite time-series, ignoring spatial context. We present a deep learning framework that maps 38 years (1984–2022) of forest disturbance across the contiguous United States by modeling temporal trajectories and spatial neighborhoods simultaneously. By leveraging a vision transformer architecture, our approach effectively filters noise from weak supervision signals to produce spatially coherent disturbance maps. We perform exhaustive evaluations across multiple satellites (Landsat, Sentinel-1, Sentinel-2) and temporal windows (38 years and the more recent 6 years), validating performance against a novel, manually annotated validation dataset (n=300) and independent fire perimeter dataset (n=706). The results highlight the complexity of the task: while our spatio-temporal model demonstrates high precision (up to 98.2\% for $\pm$1 year detection on MTBS and up to 71.3\% on the CONUS validation datasets, with F1-scores up to 75.8\% and 47.3\%, respectively) and effectively reduces spatial artifacts, it exhibits performance trade-offs across different disturbance regimes compared to pixel-wise baselines. Our method offers a promising foundation for consistent forest monitoring.
\end{abstract}

\begin{IEEEkeywords}
	Forest Disturbance Mapping, Deep Learning, Landsat.
\end{IEEEkeywords}

\section{Introduction}
\noindent Forests are critical components of the Earth's ecosystem, regulating the global carbon cycle, sustaining biodiversity, and providing other essential ecosystem services~\cite{brockerhoff2017forest}. However, forest landscapes are increasingly threatened by a complex combination of anthropogenic pressures, such as logging and agricultural conversion, and climate-induced hazards, such as wildfires and insect outbreaks~\cite{ferrer2019ecosystem}. To manage forest resources effectively, accurate and large-scale monitoring is required. Satellite remote sensing has become one of the standard tools for this task~\cite{fassnacht2024remote}. In combination with machine learning methods, it enables the systematic mapping of key forest attributes, such as forest extent, typology, canopy height, and biomass stocks over large areas~\cite{potapov2008mapping, neumann2025natural, lang2023high, sialelli2024agbd}.

\begin{figure}
    \centering
    \includegraphics[width=1\linewidth]{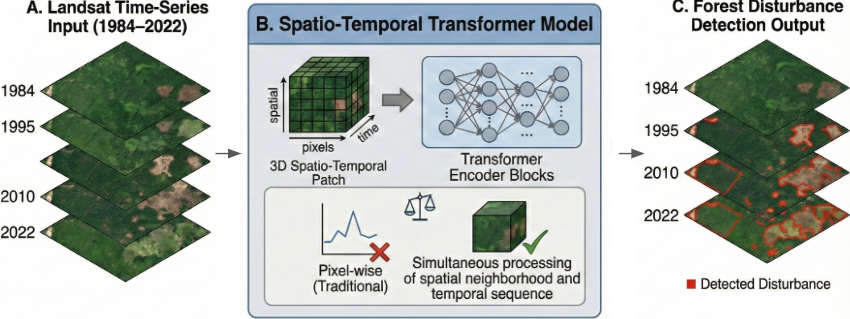}
    \caption{We propose a method that utilizes Landsat satellite image time-series in a deep learning framework to detect forest disturbances across the contiguous United States.}
    \label{fig:placeholder}
\vspace*{-2em}
\end{figure}

\noindent Despite the availability of dense temporal records, such as the Landsat satellite remote sensing data archive~\cite{wulder2016global}, accurate disturbance detection remains a significant methodological challenge. Current operational approaches typically rely on pixel-based time-series analysis (e.g., LandTrendr~\cite{kennedy2010detecting}, CCDC~\cite{zhu2014continuous}, BFAST~\cite{verbesselt2010detecting}) that analyze the spectral history of individual pixels in isolation. While these pixel-wise methods are effective at detecting both gradual and abrupt spectral changes, they inherently lack spatial context. Pixel-wise approaches treat spatial events like logging or wildfires as a collection of independent data points rather than contiguous spatial phenomena. This limits their ability to accurately model such events and can introduce spatial inconsistencies in their predictions. Noisy input data---caused by atmospheric artifacts or geometric registration errors---pose a particular challenge for pixel-wise methods, often necessitating complex ad-hoc post-processing to achieve spatial consistency. In addition, information from high-frequency patterns and texture is not available to these methods, a limitation that becomes more severe as remote sensing data resolution increases. To improve the reliability of continental-scale forest monitoring, this work explores moving beyond isolated pixel analysis to map historical forest disturbances. We propose a framework that simultaneously models the temporal evolution and spatial geometry of forest disturbances in an end-to-end deep learning setup.
The contributions of our work are as follows: \textbf{(1)} We propose a novel spatio-temporal transformer-based approach for forest disturbance detection from Landsat time-series. \textbf{(2)} We introduce a new, manually labeled validation dataset for forest disturbance detection in the contiguous United States. \textbf{(3)} We systematically evaluate the estimated forest disturbances against state-of-the-art disturbance detection methods.
\begin{figure}[b]
    \centering
    \includegraphics[width=1\linewidth]{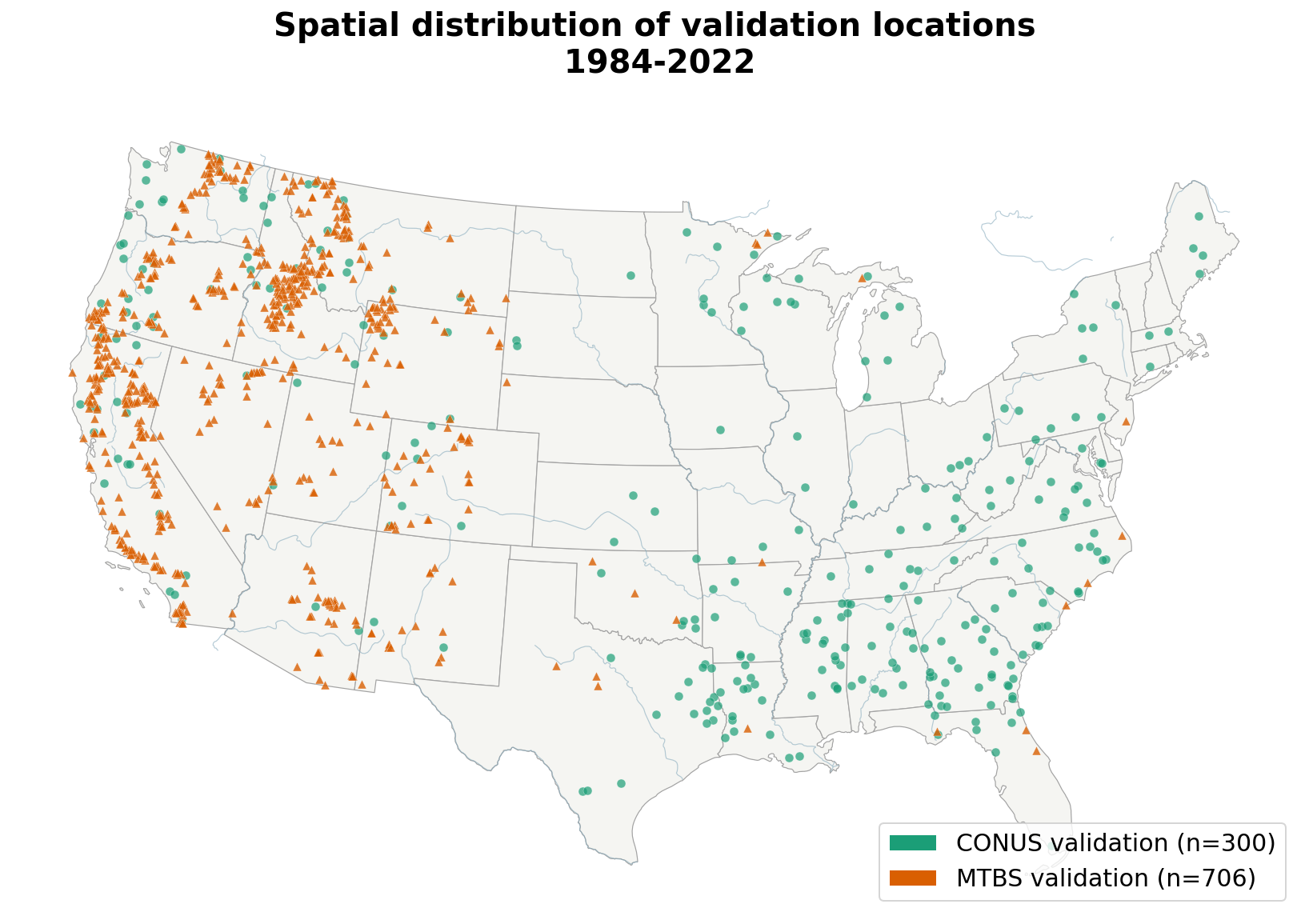}
    \caption{Locations of the CONUS validation and MTBS validation sites with forest disturbance annotations from 1985-2022 in the contiguous United States.}
    \label{fig:val_locations}
\end{figure}
\begin{figure*}
    \centering
    \includegraphics[width=1\textwidth]{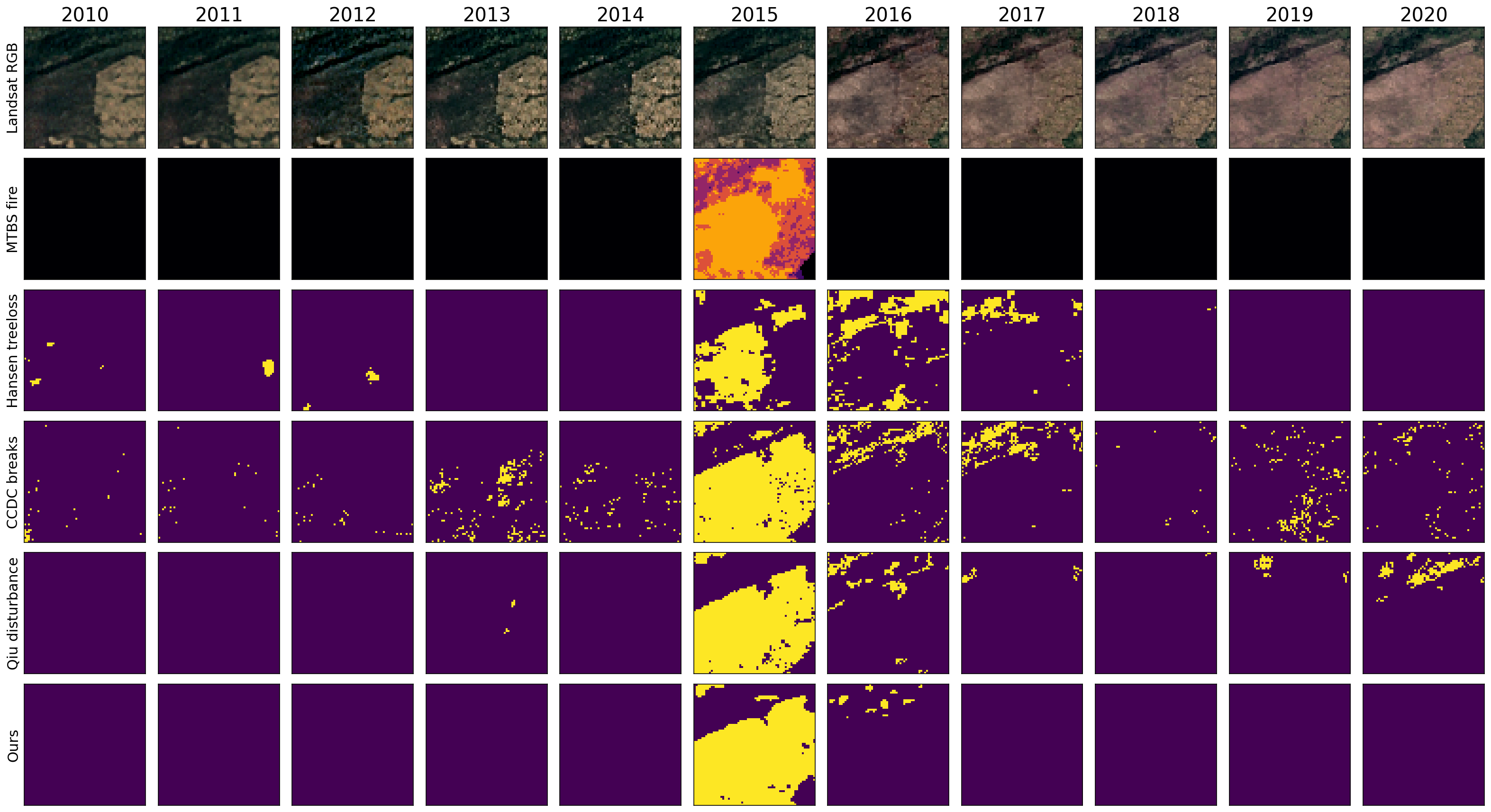}
    \caption{Example of a Landsat time-series (top row) for a location that was disturbed by fire in 2015 (MTBS fire intensity, second row~\cite{finco2012monitoring}). Subsequent rows show binary disturbance annotations from different sources: Hansen treeloss~\cite{hansen2013high}, CCDC~\cite{zhu2014continuous} and Qiu disturbances~\cite{qiu2025shift}. Predictions from our proposed method are shown in the bottom row and exhibit fewer missing detections in disturbed areas, as well as fewer false positive detections at individual pixels.}
    \label{fig:predictions}
\end{figure*}

\section{Related Work}
\noindent Remote sensing data provide a scalable solution for accurate forest monitoring~\cite{fassnacht2024remote}. In particular, the Landsat archive offers a multi-decadal record of global multi-spectral satellite remote sensing observations at 30-meter spatial resolution and has become a standard source for large-scale historical analyses~\cite{wulder2016global}. While early forest mapping relied on parametric statistics, the field has shifted toward machine learning with non-parametric methods like random forests and deep learning architectures. These approaches are now widely used for forest mapping, typology classification, and biomass estimation~\cite{waldeland2022forest, neumann2025natural, sialelli2024agbd}.\\
Landsat data has been used to produce high-resolution (30m) global maps of forest cover loss and gain, providing a consistent baseline for tracking deforestation~\cite{hansen2013high}. Subsequent work additionally classified the drivers of this loss (e.g., commodity-driven deforestation vs. shifting agriculture)~\cite{curtis2018classifying, sims2025:erl:drivers} and quantified changes in the frequency of different disturbance drivers across areas of interest~\cite{qiu2025shift} at the country scale.

Despite these advances, accurate disturbance detection remains challenging, as it is typically addressed with pixel-wise methods such as LandTrendr, CCDC, or BFAST~\cite{kennedy2010detecting, zhu2014continuous, verbesselt2010detecting} that analyze individual pixels in isolation~\cite{paulino2024forest}. Methods like convolutional neural networks and vision transformers can model spatio-temporal relationships simultaneously, albeit at higher computational cost~\cite{ronneberger2015u,dosovitskiyimage}. Existing deep learning methods focus on near-real-time detection~\cite{du2023combined}, processing of very high-resolution imagery~\cite{kislov2021extending, hamdi2019forest}, or attributing disturbance drivers~\cite{chen2021mapping,cue2025deep}.


\section{Methodology}
\subsection{Datasets}
\noindent\textbf{Landsat}: Satellites from the Landsat family have provided continuous global remote sensing data for over 50 years. We utilize Landsat 5, 7, 8 and 9 to obtain annual mosaics of the contiguous United States at 30m spatial resolution with 6 spectral channels (RGB, near-, and short-wave infrared bands). We create annual mosaics for 1985-2022, apply cloud-filtering, mask invalid pixels, and generate composites by computing the median of the remaining observations.\\
\noindent\textbf{Sentinel-1/2}: In addition to the long-term Landsat dataset, we also utilize data from the Copernicus program to estimate forest disturbances for the 2017-2022 time range. Sentinel-2 provides optical multi-spectral observations at 10-60m resolution (the 60m channels were not used in this work). We also use synthetic-aperture radar (SAR) data from Sentinel-1 at C-band wavelength that is insensitive to clouds. Specifically, we use VV/VH polarizations from ascending orbits, which provide better coverage for our area of interest. The data are resampled to 30m spatial resolution for comparability with Landsat, and processed into annual composites.\\
\textbf{Weak Training Labels}:
To create a large-scale supervised training dataset for forest disturbance detection, we utilize disturbance detections in the contiguous United States from prior work~\cite{qiu2025shift} as weak labels. Disturbance data is extracted for each year from 1985 to 2022 and processed into binary segmentation masks. We create two separate datasets: \texttt{planted\_34k}, with \num{34000} locations sampled across planted forests in the Southeast of the United States, and \texttt{conus\_100k}, a random stratified sample of \num{100000} locations across the entire contiguous United States. Stratification is based on forest typology~\cite{neumann2025annual}: 33\% of locations in natural forest, 34\% in planted forest, and 33\% on other land. Both datasets are split with an 8:1:1 ratio into training, validation and test datasets.\\
\textbf{CONUS validation dataset}: To independently evaluate the accuracy of forest disturbance detections we manually annotate a reference dataset of disturbance events. We create a stratified random sample of 300 locations across the contiguous United States with more than 10\% treecover based on~\cite{hansen2013high} (see Fig.~\ref{fig:val_locations}). The dataset contains 100 locations with no disturbance in 1985-2022, 100 locations with one disturbance, and 100 locations with multiple disturbances (based on~\cite{qiu2025shift}). For each location, we create binary per-image annotations (\textit{disturbance} or \textit{no disturbance}) for every year from 1985-2022, which yields a total of \num{11400} annotations. Data is annotated in Google EarthEngine~\cite{gorelick2017google}, using static Landsat RGB data for all years and time-series of the normalized burn ratio~\cite{escuin2008fire}, similar to the TimeSync tool~\cite{cohen2010detecting}.\\
\textbf{MTBS validation dataset}: As additional validation data we utilize wildfire footprints from the Monitoring Trends in Burn Severity (MTBS) program~\cite{finco2012monitoring}. This data contains spatially explicit information about historical wildfires in the United States. The wildfires are categorized by intensity (low, moderate, high) and we consider high-intensity fires as disturbance events. Validation locations are randomly sampled across the United States and we retain 706 point locations that fall within the perimeter of high-intensity fires according to MTBS between 1985-2022, and contain 10\% or more forest cover according to~\cite{hansen2013high} (see Fig.~\ref{fig:val_locations}).

\begin{table*}[t]
    \centering
    \caption{Performance on CONUS and MTBS validation tasks of baselines and our approaches with varying input modalities\\ (using \texttt{conus\_100k} (C100k) and \texttt{planted\_34k} (P34k) training datasets, and denoting $\pm$standard deviation).}
    \label{tab:combined}
    \footnotesize
    
    \newcolumntype{Z}{>{\centering\arraybackslash}X}
    
    \setlength{\tabcolsep}{2.5pt} 
    \renewcommand{\arraystretch}{1.1} 
    
    \begin{tabularx}{\textwidth}{@{} ll | ZZZZ | ZZZZ ll@{}}
    \toprule
    \multicolumn{2}{c|}{} & \multicolumn{4}{c|}{\textbf{CONUS Validation}} & \multicolumn{4}{c}{\textbf{MTBS Validation}} \\
    
    \textbf{Years} & \textbf{Approach/Inputs} &  \textbf{F1} & \textbf{Recall} & \textbf{Prec} & \textbf{Prec$\pm$1y} & \textbf{F1} & \textbf{Recall} & \textbf{Prec} & \textbf{Prec$\pm$1y} \\
    \midrule

    \multirow{5}{*}{\shortstack{1985--\\2022}} & CCDC~\cite{zhu2014continuous} 
        & 30.68 & 44.35 & 23.45 & 36.06 
        & 52.65 & \textbf{69.28} & 42.45 & 52.89 \\
    & Qiu~\cite{qiu2025shift}  
        & 39.72 & \textbf{48.12} & 33.82 & 49.71 
        & {62.46} & 61.89 & {63.04} & 76.83 \\
    \noalign{\smallskip} \cdashline{2-10} \noalign{\smallskip}
    & Landsat (P34k)
        & \textbf{44.9}\tiny$\pm$1.3 & 40.5\tiny$\pm$1.7 & \textbf{50.5}\tiny$\pm$1.3 & \textbf{70.7}\tiny$\pm$1.3 
        & 37.0\tiny$\pm$3.5 & 27.2\tiny$\pm$3.6 & 58.5\tiny$\pm$1.6 & 91.4\tiny$\pm$1.1 & \ldelim\}{2}{2.5pt} & \multirow{2}{*}{\rotatebox{90}{Ours}} \\
    & Landsat (C100k)
        & {43.0}\tiny$\pm$0.8 & 39.5\tiny$\pm$0.9 & {47.3}\tiny$\pm$1.3 & {70.1}\tiny$\pm$1.7 
        & \textbf{68.0}\tiny$\pm$0.7 & 57.0\tiny$\pm$0.9 & \textbf{84.2}\tiny$\pm$0.3 & \textbf{92.5}\tiny$\pm$0.6 \\

    \midrule
    
    \multirow{14}{*}{\shortstack{2017--\\2022}} & CCDC~\cite{zhu2014continuous}
        & 23.85 & 40.63 & 16.88 & 33.77 
        & 59.83 & \textbf{72.27} & 51.04 & 62.91 \\
    & Qiu~\cite{qiu2025shift}
        & 31.71 & 40.63 & 26.00 & 48.00 
        & {69.34} & 68.91 & 69.79 & 82.55 \\
    & Hansen~\cite{hansen2013high}
        & 38.60 & 34.38 & 44.00 & {68.00} 
        & 65.71 & 57.56 & 76.54 & 95.53 \\

    \noalign{\smallskip} \cdashline{2-10} \noalign{\smallskip}

    & Landsat (P34k)
        & \textbf{47.3}\tiny$\pm$3.1 & \textbf{46.9}\tiny$\pm$4.4 & \textbf{47.9}\tiny$\pm$2.6 & 65.0\tiny$\pm$2.1 
        & 38.0\tiny$\pm$2.3 & 27.9\tiny$\pm$2.0 & 59.6\tiny$\pm$2.8 & 91.0\tiny$\pm$0.1 & \ldelim\}{10}{2.5pt} & \multirow{10}{*}{\rotatebox{90}{Ours}} \\
    & Sentinel-1 (P34k)
        & 27.6\tiny$\pm$5.3 & 22.9\tiny$\pm$3.9 & 34.8\tiny$\pm$7.8 & 64.3\tiny$\pm$4.1 
        & 45.0\tiny$\pm$1.5 & 29.6\tiny$\pm$1.0 & \textbf{94.2}\tiny$\pm$3.1 & \textbf{98.2}\tiny$\pm$2.6 \\
    & Sentinel-2 (P34k)
        & 41.3\tiny$\pm$0.6 & 39.6\tiny$\pm$1.5 & 43.5\tiny$\pm$3.2 & 62.6\tiny$\pm$3.2 
        & 45.2\tiny$\pm$3.6 & 35.6\tiny$\pm$2.8 & 62.3\tiny$\pm$7.4 & 93.2\tiny$\pm$2.0 \\
    & Sentinel-2 + S1 (P34k)
        & 38.1\tiny$\pm$3.0 & 36.5\tiny$\pm$5.3 & 40.3\tiny$\pm$1.6 & 60.1\tiny$\pm$3.8
        & 54.0\tiny$\pm$3.3 & 44.5\tiny$\pm$2.7 & 68.5\tiny$\pm$4.7 & 92.3\tiny$\pm$ 1.8 \\
    & Landsat + S1 (P34k)
        & 37.8\tiny$\pm$5.0 & 36.5\tiny$\pm$6.4 & 39.5\tiny$\pm$3.2 & 56.5\tiny$\pm$3.8
        & 42.6\tiny$\pm$1.4 & 31.0\tiny$\pm$1.8 & 69.2\tiny$\pm$5.3 & 93.4\tiny$\pm$1.0 \\


    
    

    & Landsat (C100k)
        & 36.3\tiny$\pm$1.8 & 36.5\tiny$\pm$1.5 & 36.1\tiny$\pm$2.2 & 68.1\tiny$\pm$2.0 
        & 71.0\tiny$\pm$1.1 & 60.8\tiny$\pm$0.9 & 85.3\tiny$\pm$2.3 & 96.9\tiny$\pm$0.2 \\
    & Sentinel-1 (C100k)
        & 35.6\tiny$\pm$1.7 & 33.3\tiny$\pm$1.5 & 38.1\tiny$\pm$2.1 & 66.6\tiny$\pm$4.2 
        & 66.6\tiny$\pm$0.6 & 51.7\tiny$\pm$0.6 & {93.7}\tiny$\pm$0.4 & {97.7}\tiny$\pm$0.6 \\
    & Sentinel-2 (C100k)
        & 37.2\tiny$\pm$1.3 & 36.5\tiny$\pm$1.5 & 38.1\tiny$\pm$1.4 & 69.7\tiny$\pm$3.2 
        & 75.3\tiny$\pm$3.3 & 66.7\tiny$\pm$4.0 & 86.5\tiny$\pm$2.5 & 95.6\tiny$\pm$1.1 \\
    & Sentinel-2 + S1 (C100k)
        & 37.6\tiny$\pm$7.5 & 35.4\tiny$\pm$10.3 & 41.9\tiny$\pm$3.9 & \textbf{71.3}\tiny$\pm$3.6 
        & \textbf{75.8}\tiny$\pm$3.5 & 65.4\tiny$\pm$7.0 & 91.3\tiny$\pm$4.3 & 95.3\tiny$\pm$4.5  
        \\
    & Landsat + S1 (C100k)
        & 36.0\tiny$\pm$3.5 & 36.5\tiny$\pm$3.9 & 35.7\tiny$\pm$3.1 & 68.4\tiny$\pm$0.5
        & 72.9\tiny$\pm$0.8 & 62.3\tiny$\pm$1.1 & 87.8\tiny$\pm$0.2 & 95.5\tiny$\pm$0.7 
        \\
    \bottomrule
    \end{tabularx}
    \vspace{-1.5em}
\end{table*}

\subsection{Model \& Training}
\label{sec:methods}
\noindent In this work we adapt the Multi-modal Temporal Spatial Vision Transformer (MTSViT) model~\cite{jiang2025not} to detect forest disturbances in satellite image time-series. The original MTSViT model is designed for forest mapping from multi-modal satellite image time-series. The main components are a spatial encoder, a temporal encoder, a multi-modal decoder, and a multi-temporal segmentation head. The attention mechanism~\cite{vaswani2017attention} is used to exploit long temporal and spatial context. Encoders and the decoder consist of two transformer layers and all modalities are processed independently in the spatial and temporal encoders. We train the model $f_\theta$ to predict disturbance label annotations $y$ for all years simultaneously in one step from the time-series input $x$:
\begin{equation}
        \hat{y}=f_\theta(x), \qquad \text{with }x \in \mathbb{R}^{t,h,w,c}, \qquad \text{and }\hat{y} \in \mathbb{R}^{t,h,w}
\end{equation}
In practice we choose a spatial size of 1920m per image chip ($h,w=64,64$ pixel at 30m resolution) and a temporal sequence length of 38 years (1985-2022). The model is trained with the Adam optimizer~\cite{kingma2014adam} and binary cross-entropy loss. A standard threshold of $0.5$ is applied to create binary disturbance classifications. We report model performance with F1-score, precision and recall metrics. Additionally, we define a temporally-tolerant precision metric, "Prec~$\pm 1$y", which considers a disturbance detection as correct if it occurred within $\pm 1$~years of an annotated disturbance event. This metric accounts for temporal uncertainty in the labeling (\textit{ground-truth}) of the annotated data.
We report model performance along with standard deviations computed from three independent training runs.

\section{Results}
We train the MTSViT model separately on the \texttt{planted\_34k} and \texttt{conus\_100k} datasets as outlined in Sec.~\ref{sec:methods}. To evaluate model performance at forest disturbance detection, we perform inference on the independent CONUS and MTBS validation datasets. Quantitatively, we find that our method achieves strong performance relative to the long term baselines by Zhu~\cite{zhu2014continuous} (CCDC) and Qiu~\cite{qiu2025shift}. On the CONUS validation dataset, the model trained on the \texttt{planted\_34k} dataset achieves an increase in F1-score of +5\% (absolute) over Qiu et al., with a $\pm$1-year precision of more than 70\% (see Tab.~\ref{tab:combined}). We find that the geographic distribution of training data has a marked impact on model performance. When evaluated on the MTBS validation dataset (which mainly covers the Western US, see Fig.~\ref{fig:val_locations}), the model trained on \texttt{planted\_34k} (mostly sampled in the Southeast US) under-performs the baselines significantly. However, training on the more geographically balanced \texttt{conus\_100k} dataset mitigates this issue. The model achieves an increase in F1-score of +6\% over the method by Qiu et al., with a $\pm$1-year precision of 92.5\% while maintaining similar performance to the \texttt{planted\_34k} model on the CONUS validation dataset.
This performance gap between the models trained on \texttt{planted\_34k} and \texttt{conus\_100k} confirms that regional environmental factors heavily influence model performance. The disturbance regimes in the Southeast (often dominated by intensive forestry management and smaller spatial scales) do not transfer well to the large-scale wildfires prevalent in the Western US covered by the MTBS dataset. Consequently, the robustness of our method at a continental scale is not an inherent property of the architecture, but depends on using a geographically representative training dataset like \texttt{conus\_100k} to cover diverse disturbance regimes.\\
We also evaluate performance on a shorter time-frame (2017--2022). The proposed model trained on \texttt{planted\_34k} improves the F1-score by +9\% over Hansen~\cite{hansen2013high} on the CONUS validation dataset, while our \texttt{conus\_100k} model reaches the highest F1-score among all methods on the MTBS dataset at 72\% (see Tab.~\ref{tab:combined}). The 2017--2022 time-span also enables evaluation with Sentinel-1 and Sentinel-2 data. We find that optical Sentinel-2 imagery yields higher accuracy than Sentinel-1 radar data, and combining both modalities from \texttt{conus\_100k} locations achieves the highest overall F1-score of 75.8\%, an improvement of +4.5\% over Landsat.\\
Qualitatively, we find that our proposed approach yields cleaner segmentation results with consistent predictions and less pixel-wise noise or wrong classifications of individual pixels (see Fig.~\ref{fig:predictions}). In addition, we also observe fewer erroneous detections for the same location in consecutive years (i.e., double counting of disturbance events).
This ability to produce cleaner maps than the weak training labels themselves highlights a key advantage of the spatio-temporal attention mechanism. While the pixel-wise methods used to generate the weak training labels often suffer from atmospheric noise and geometric misregistrations (leading to isolated false positives), the vision transformer learns to recognize the coherent spatial and temporal signatures of actual disturbances. By aggregating context across both space and time, the model effectively filters out the uncorrelated noise present in the weak supervision signal, learning the underlying phenomenon rather than simply memorizing the noisy labels.


\section{Conclusion}
This work demonstrates that spatio-temporal modeling can improve long-term forest disturbance monitoring from satellite image time-series. By simultaneously modeling temporal trajectories and spatial neighborhoods, our approach addresses a fundamental limitation of traditional pixel-wise temporal segmentation methods. Our results indicate that this yields significant quantitative improvements in detection accuracy over baseline techniques. Furthermore, qualitative assessment confirms that the spatio-temporal model produces cleaner disturbance maps, characterized by a reduction in potentially noisy single-pixel detections and more contiguous disturbance footprints.\\
We note that the approach is sensitive to geographic distribution shift, which can be mitigated by informed training data selection. Consequently, our framework offers a robust and scalable solution for generating consistent, high-quality disturbance data necessary for large-scale ecosystem assessment and management. 

While our work successfully detects disturbances, an interesting avenue is to connect it with the attribution of specific disturbance drivers (e.g. \cite{sims2025:erl:drivers}). Future work can focus on multi-class classification to identify the drivers and extending the framework to estimate forest age and rotation times. These estimates will provide important cues for long-term carbon modeling and forest types mapping.



\clearpage
\small
\bibliographystyle{IEEEtranN}
\bibliography{references}

\end{document}